\documentclass[letterpaper, 10 pt, conference]{ieeeconf}  

\IEEEoverridecommandlockouts                              

\usepackage{amsmath,amssymb,amsfonts}
\usepackage{algorithmic}
\usepackage{textcomp}
\usepackage{xcolor}
\usepackage[textsize=footnotesize]{todonotes}
\usepackage{verbatim}
\usepackage[nolist]{acronym}
\usepackage{dirtytalk}
\usepackage{graphicx}
\usepackage{comment}
\usepackage{algorithm} 
\usepackage[footnotesize]{subfigure}
\usepackage{morefloats}
\usepackage{soul}

\newacro{vr}[VR]{Virtual Reality}
\newacro{hmd}[HMD]{Head-Mounted Display}
\newacro{ems}[EMS]{Electrical Muscle Stimulation}
\newacro{imu}[IMU]{Inertial Measurement Unit}
\newacro{bci}[BCI]{Brain-Computer Interface}
\newacro{vwg}[VWG]{Virtual World Generator}
\newacro{moo}[MOO]{Multi-Objective Optimization}
\newacro{dof}[DoF]{Degree of Freedom}
\newacro{ssq}[SSQ]{Simulator Sickness Questionnaire}
\newacro{ddr}[DDR]{Differential Drive Robot}
\newacro{fov}[FOV]{Field-of-View}

\usepackage{graphics} 
\usepackage{epsfig} 

\title{\LARGE \bf
Analysis of User Preferences for Robot Motions \\ in Immersive Telepresence}

\author{Katherine J. Mimnaugh$^{1}$, Markku Suomalainen$^{1}$, Israel Becerra$^{2}$, Eliezer Lozano$^{2}$, \\ Rafael Murrieta-Cid$^{2}$, and Steven M. LaValle$^{1}$
\thanks{This work was supported in part by the Business Finland Project HUMORcc 6926/31/2018, in part by the Academy of Finland Project PERCEPT, 322637, in part by the US National Science Foundation under Grants 035345 and 1328018, and in part by the Secretaría de Innovación, Ciencia Y Educación Superior SICES under Grant SICES/CONV/250/2019 CIMAT. (Corresponding author: Katherine Mimnaugh.)}
\thanks{$^{1}$Katherine Mimnaugh, Markku Suomalainen, and Steven LaValle are with the Center of Ubiquitous Computing, Faculty of Information Technology and Electrical Engineering, University of Oulu, Finland {\tt\small (e-mail: firstname.surname@oulu.fi).}}%
\thanks{$^{2}$Israel Becerra, Eliezer Lozano, and Rafael Murrieta-Cid are with the Centro de Investigación en Matemáticas (CIMAT), Guanajuato, Mexico {\tt\small(e-mail: israelb@cimat.mx; eliezer.lozano@cimat.mx; murrieta@cimat.mx).}}%
}

\begin{document}

\maketitle

\thispagestyle{empty}
\pagestyle{empty}

\begin{abstract}
This paper considers how the motions of a telepresence robot moving autonomously affect a person immersed in the robot through a head-mounted display. In particular, we explore the preference, comfort, and naturalness of elements of piecewise linear paths compared to the same elements on a smooth path. In a user study, thirty-six subjects watched panoramic videos of three different paths through a simulated museum in virtual reality and responded to questionnaires regarding each path. Preference for a particular path was influenced the most by comfort, forward speed, and characteristics of the turns. Preference was also strongly associated with the users' perceived naturalness, which was primarily determined by the ability to see salient objects, the distance to the walls and objects, as well as the turns. Participants favored the paths that had a one meter per second forward speed and rated the path with the least amount of turns as the most comfortable. 
\end{abstract}

\section{INTRODUCTION}\label{sec:intro}

Immersive robotic telepresence enables people to embody a robot in a remote location, such that they can move around and feel as if they were really there instead of the robot. Currently, the most scalable technology with the potential to achieve the feeling of being in a remote location is a \ac{vr} \ac{hmd} with a 360$^{\circ}$ camera, or a limited \ac{fov} camera attached to a very fast pan-tilt unit on board a mobile robot. The immersion created by the \ac{hmd}, combined with the user's ability to look around at the robot's environment by simply rotating their head, has the potential to create a feeling of \textit{presence}. Presence, the feeling of \textit{being there}, is a term commonly used in \ac{vr} research \cite{sanchez2005presence} to describe an important aspect of the immersive experience. The immersion provided by an \ac{hmd} has been shown to increase presence \cite{banos2004immersion}, which can facilitate more natural interaction and thus ease the difference between telepresence and physical presence \cite{slater1997framework}. 

\begin{figure}
\centering
\includegraphics[width=\columnwidth]{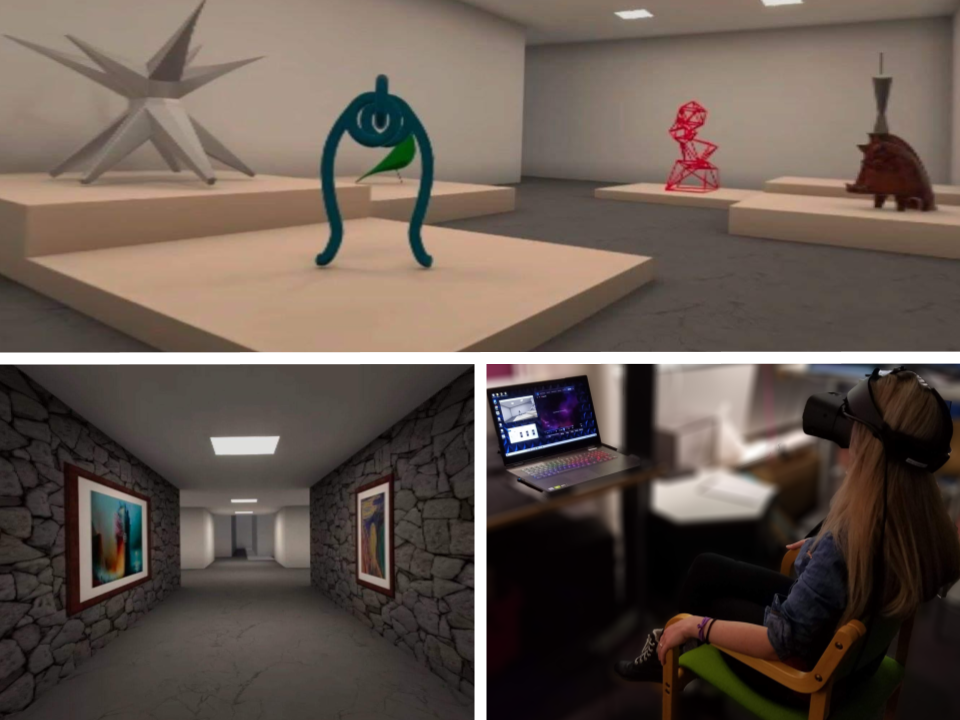}
\caption{Top: a screenshot from inside the gallery of the virtual museum, seen by the participant from the point of view of the virtual robot. Bottom left: a screenshot of the hallway entrance in the virtual museum. Bottom right: a participant watching a video of a path in VR during the user study. 
}
\label{fig:teleop} 
\vspace{-5mm}
\end{figure}

Unfortunately, there are several concerns regarding the use of telepresence with an \ac{hmd} that do not typically occur in telepresence with a Fixed Naked-eye Display (FND, a traditional display such as on a computer or a phone). The velocities that the robot uses or the passing distances to objects and walls, which would otherwise be comfortable in FND telepresence, can be uncomfortable in immersive telepresence. Second, a serious consequence of using an \ac{hmd} is \textit{VR sickness} \cite{LaValle_bookVR}, which can result in the user experiencing discomfort, nausea, or vertigo. Although there are known methods to reduce VR sickness \cite{Chang_Kim_Yoo_2020, Teixeira_Palmisano_2020}, 
these techniques may have a negative effect on presence \cite{weech2019presence}. Such issues can be overcome by planning autonomous trajectories for the robot that avoid motions known for contributing to VR sickness. For example, turns have been shown to play a role in VR sickness \cite{hu1999systematic}, but there is evidence that the duration of the turn is more important than the total angle of rotation \cite{Widdowson_Becerra_Merrill_Wang_LaValle_2019}. We used this knowledge to plan piecewise linear trajectories for a robot, consisting of straight lines in 2-D space and rotations in place, which 
reduced VR sickness in an immersive telepresence experience \cite{becerra2020human}. 

However, VR sickness is not the only concern; issues such as the path not feeling natural or qualities of the turns when rotating in place may deteriorate the experience. These issues are typically not a problem in telepresence through an FND, since it is the immersion experienced through the near-eye displays of an \ac{hmd} that allows the user to really feel them. These issues are also not frequently studied in pure VR research, since motions in VR are usually either directly controlled by the user, or teleportation is used. Although these methods for locomotion in VR may be more comfortable for the user, teleportation is infeasible in telepresence and manual control is perceived as bothersome in large environments \cite{rae2017robotic}. Thus, there is a need to research autonomous motions to overcome these obstacles in immersive telepresence.

Another interesting aspect regarding the telepresence experience is the \textit{naturalness} of the robot's motions. There has been discussion on the importance for robots to move naturally \cite{laviers2019make}, and several works have attempted to make robots perform more natural motions \cite{kwon2008natural, Turnwald_Wollherr_2019}. However, to the authors' knowledge, there is no widely accepted definition of natural motions for mobile robots. Moreover, from the perspective of the user embodying the telepresence robot, there does not appear be any research on whether people even prefer motions that they would perceive as natural, or more generally what kind of trajectories are preferred for an autonomously moving immersive telepresence robot to take. 

We present the first results on how users experience autonomous motions when immersed in a telepresence robot through an HMD. In particular, we consider the interplay between preferences, comfort, and the perceived naturalness of motions of an autonomously moving telepresence robot from a first-person perspective. We performed an experiment where users wore an \ac{hmd} and watched three pre-recorded trajectories of a telepresence robot moving through a virtual museum. Two of the trajectories were piecewise linear trajectories with 45$^{\circ}$ or 90$^{\circ}$ turns, optimized for the least amount of turns and for the shortest path \cite{becerra2020human}, while the third path was a smooth path created with a Rapidly-exploring Random Trees (RRT) algorithm \cite{lavalle2001randomized}. We selected the RRT because it is a commonly used algorithm for robot motion planning, it can plan trajectories in 5-D space while accounting for the system dynamics of a \ac{ddr}, and because there is evidence that RRT paths are perceived as human-like \cite{Turnwald_Wollherr_2019}. 

We found that comfort had a strong effect on path preference, and that the subjective feeling of naturalness also had a strong effect on path preference, although people consider different things as natural. We examined the answers to open-ended questions about naturalness and found noteworthy variation in how subjects thought of naturalness in terms of the time frame, and either in regards to the specific context or more generally. When selecting which path was preferred, most comfortable, and most natural, the turns were one of the top three reasons mentioned for each. Finally, we also found that the robot's speed and passing distances to objects had a significant impact on the users and must be carefully considered when designing for immersive telepresence. 
 

\section{RELATED WORK}

There is an increasing amount of work on telepresence with an HMD, with many regarding physical object manipulation \cite{hetrick2020comparing,whitney2020comparing}. Oh et al. \cite{oh2018360} used a robot on a virtual tour. Stotko et al. \cite{stotko2019vr} created a virtual version of the environment while moving. Garc{\'\i}a et al. \cite{garcia2015natural} performed experiments in a simulated underwater scenario and found that users performed better with an HMD than using a monitor. Martins and Ventura \cite{martins2009immersive} compared an HMD to a computer screen in a search-and-rescue task and concluded that the HMD provided superior situational awareness. In a leisure time comparison between a 360$^{\circ}$ camera and a wide \ac{fov} camera, it was reported that users enjoyed the panoramic experience more due to the freedom of being able to look around \cite{heshmat2018geocaching}. When considering control sharing in HMD-based telepresence waypoint navigation, choosing a destination within visible range and letting the robot move there autonomously was found to be preferred over joystick usage \cite{baker2020towards}. Together with the complaints about joysticks in other work \cite{rae2017robotic}, it is apparent that development of autonomous navigation for telepresence robots is merited. 

However, there is a limited amount of research regarding how the autonomous motions should be generated for HMD-based telepresence. When considering how a robot moves, the concept of \textit{naturalness} is often mentioned positively, to which several works have given different definitions: simply \say{human-likeness} \cite{kwon2008natural}, which may reflect organic optimization \cite{Arechavaleta_Laumond_Hicheur_Berthoz_2008}, as is used often for walking robots \cite{arakawa1997natural}; or smoothness \cite{sisbot2010synthesizing} or variability of trajectories \cite{kretzschmar2016socially}, as is used often for mobile wheeled robots. Conversely, the uncanny valley effect, originally describing how robots cause discomfort when the human-likeness of their appearance is very high \cite{mori2012uncanny}, has been tested on the human-like motions of synthetic agents with contradicting results \cite{thompson2011perception}. Nevertheless, to the knowledge of the authors, all currently available work regarding naturalness focuses on how the robot's motions are perceived from the outside. In contrast, in this work we explore what sort of trajectories a user immersed inside of a telepresence robot considers natural. Moreover, we evaluate whether perceived naturalness affects the comfort or preference of the user aboard the telepresence robot.

\section{METHODS}\label{sec:methods}
\subsection{Participants}
Forty-five individuals from the University of Oulu campus and local community were recruited through flyers, Facebook posts, email lists, and by word-of-mouth. One subject was excluded based on an initial report of severe nausea, blurred vision, and dizziness before the stimuli began. Three subjects withdrew from the study before it was finished due to significant VR sickness symptoms after viewing one of the videos. Two subjects were excluded due to equipment failure resulting in a loss of head tracking during the videos. One subject was excluded due to experimenter error in administration of the questionnaires, and two subjects were excluded for not answering questions on a questionnaire. 

The final sample consisted of 36 people. Participants ranged in age from 20 to 44, with a mean age of  28.25 years. The group was divided evenly between women (n=18) and men (n=18). Subjects were randomly assigned into groups which were fully counterbalanced by gender and by path video display order to counteract any order effects. Ten subjects reported having never used a virtual reality headset previously (28$\%$), 15 subjects reported having used VR between one and nine times or at least a few times (42$\%$), 10 subjects reported having used VR ten times or more, frequently or on a regular basis (28$\%$), and one person did not respond regarding their previous VR use (2$\%$). Nine participants reported that they never play any video games on mobile, PC, or console (25$\%$), five participants reported rarely playing video games (14$\%$), 10 reported playing often or weekly (28$\%$), seven reported playing daily (19$\%$) and five did not respond (14$\%$).

\subsection{Setup}

\begin{figure}[t]
    \centering
    \includegraphics[width=\columnwidth]{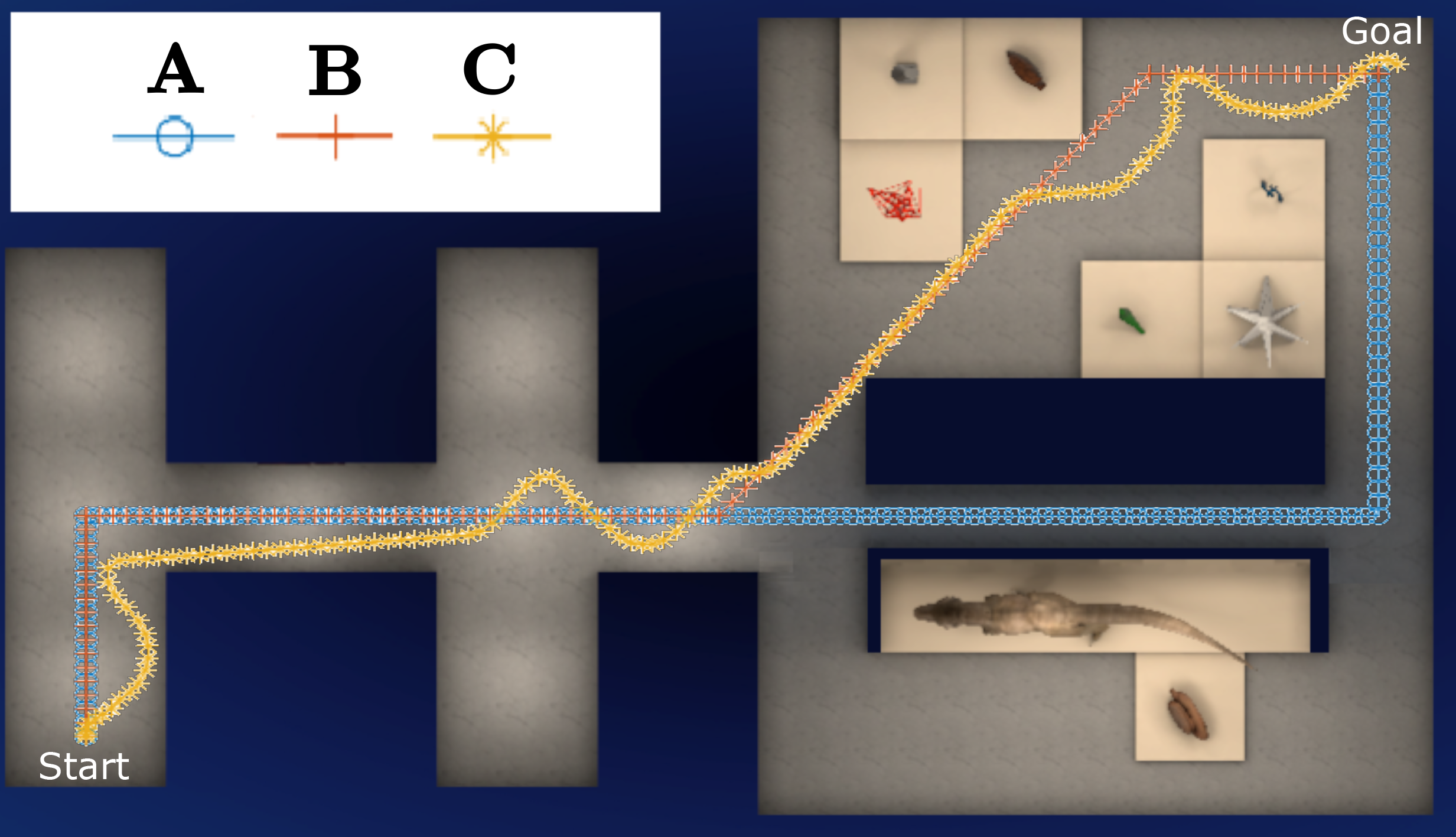} 
    \caption{A top-down view of the museum with the three tested trajectories. The Pareto optimal path that minimizes the number of turns (PLT) is labeled as {\bf A}. The Pareto optimal path that focus on reducing distance (PSP) is labeled as {\bf B}. The tested RRT path is labeled as {\bf C}.}
    \label{fig:museum} 
    \vspace{-3mm}
\end{figure}

The subjects were run on campus in our laboratory in the second week of February 2020, before COVID-19 was declared a global pandemic. The experimental environment was designed with Unity 3D and recorded as 360$^{\circ}$ videos, which were played for subjects in the Oculus Rift S headset using the Virtual Desktop application. If the paths had been computed in real time, there could have been slight differences in presentation between subjects. Therefore, to preclude the possibility that these variations could confound the results, we used pre-recorded videos. The subjects were seated, as seen in Fig.~\ref{fig:teleop}, which was taken during an experiment. Numerical details about each path can be found from Tab.~\ref{tab:path_data}, and the paths are presented in Fig.~\ref{fig:museum}.

When designing the environment, we wanted a realistic amount of optical flow through salient objects, so there were paintings and statues in the museum. The paths through the museum were different depending on the algorithm and Pareto optima (see Fig.~\ref{fig:museum}), such that the artwork was seen more closely on the Pareto Shortest Path (PSP) and the RRT than on the Pareto Least Turns (PLT) path. The PSP and PLT were piecewise linear, where rotation only happened when there was no forward motion, whereas the RRT was a smooth path, where rotation and translation occurred simultaneously. How the paths were created is described in detail in \cite{becerra2020human}. 

\subsection{Procedure}
When subjects first came into the laboratory, they were given information about the study, and signed a consent form to participate. Next, they filled out the Simulator Sickness Questionnaire (SSQ) \cite{kennedy1993simulator} for a baseline measure of sickness symptoms. Participants were then seated in a chair and given general instructions and preparation about the task. 
After watching each of the three videos, subjects removed the HMD and were handed a laptop on which to fill out a post-exposure SSQ questionnaire. Then, they were given a survey consisting of questions with ratings on a 6-point Likert scale asking about the path that they just viewed. The questions asked participants to rate how well they could find their way back to the beginning of the museum, the distance between themselves and the walls or objects, the speed when moving forward, the speed during the turns, and finally how natural (which we phrased as: ``if you consider how you would have moved through the museum if you were in control") the path through the museum was. After they completed the survey, the procedure was repeated for the other two videos based on their counterbalancing order. 

After the third video, the survey contained additional questions asking the subject to select which of the three paths they preferred, which was the most comfortable and which was the most natural. After each choice for one of the three paths, they were given the open-ended question, ``why?". The final four questions asked if they had ever used VR before, if they play video games, for their age, and for their gender. Once they had completed the final survey, they were debriefed about the purpose of the study and given a voucher for a coffee or tea from the local caf{\'e}. 

\section{RESULTS}
All analyses were exploratory and were run in SPSS with a $95\%$ confidence interval and with two-tailed significance levels at 0.05. Significance values were adjusted within each using Bonferroni correction for multiple tests. Effect sizes (Kendall's W and Cohen's w) \cite{Cohen_1988} were calculated with Psychometrica freeware \cite{lenhard2016calculation} and post hoc power analyses were calculated using G*Power \cite{Faul_Erdfelder_Lang_Buchner_2007}. For open-ended questions, the responses were categorized using thematic analysis with an inductive approach \cite{Patton2005qualitative} by two independent coders. The codes were based on the words that the subjects used in their responses, so there was some ambiguity in the specific meaning of the codes. All raw subject data, the videos used as the stimuli, and all materials presented to subjects are available at: https://osf.io/et9gy. 

This dataset was previously used in \cite{becerra2020human}, which describes the technical implementation of the paths. That paper included the distribution of responses for the question of the preferred path and the most comfortable path, and analyzed the differences in post-treatment SSQ scores after each path. The SSQ scores for the RRT were significantly higher (indicating more sickness) than those of the PLT, but not the PSP. The responses regarding naturalness, the Likert scale data, the path comparisons, and the open-ended questions which are reported below are not published elsewhere.

\subsection{Quantitative Data}



\begin{figure}[t]
\centering
\includegraphics[scale=0.43]{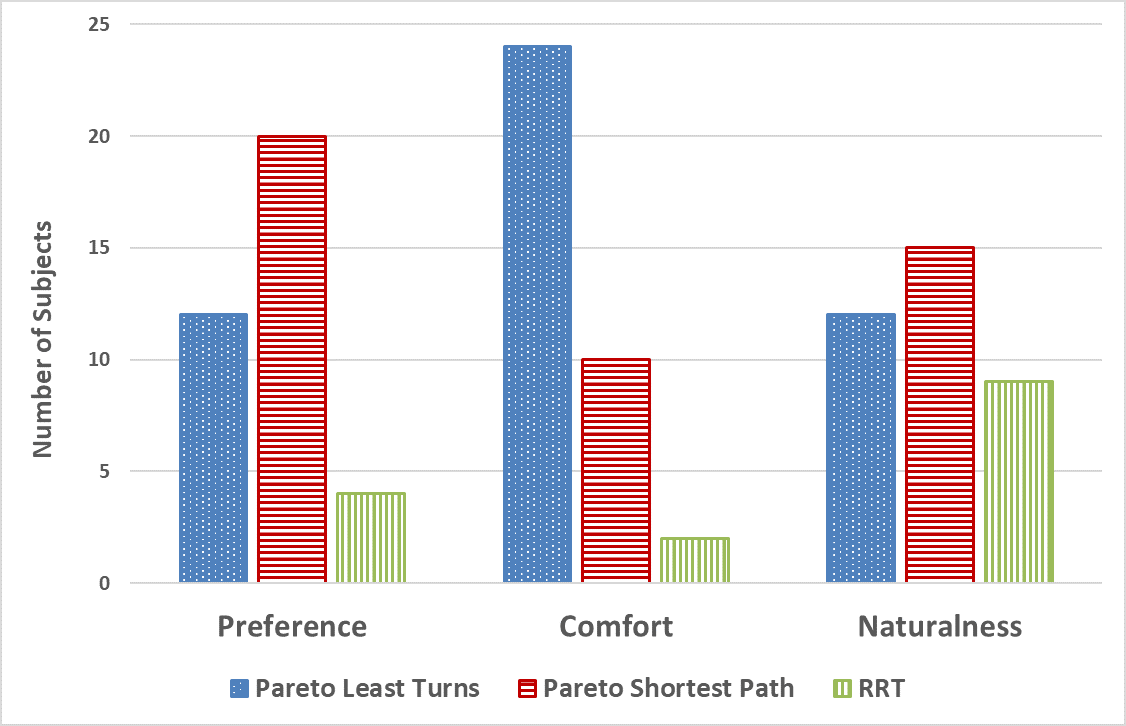}
\caption{The distribution of responses regarding which path was preferred, the most comfortable, and the most natural from a questionnaire asking users to select one of the three paths. 
}
\label{fig:chartComp} 
\vspace{-5mm}
\end{figure}

\subsubsection{PSP was preferred}
When asked to choose which of the three paths was preferred, PSP was selected the most (20 times, $56\%$), PLT was selected second most (12 times, $33\%$), and the RRT was selected the least (4 times, $11\%$).  Fig.~\ref{fig:chartComp} depicts the distribution of answers.

\subsubsection{PLT was most comfortable, and Pareto paths were more comfortable than the RRT}
When asked to select the most comfortable path, the PLT path was chosen the most at 24 times ($67\%$), PSP was selected 10 times ($28\%$), and the RRT was selected the least with only two subjects rating that path as the most comfortable ($5\%$). The Likert comfort ratings from 1 (very uncomfortable) to 6 (very comfortable) were compared across paths with a Friedman's test. There was a statistically significant difference between the PLT ($mean = 3,94, sd = 1.39$), the PSP ($mean = 3.67, sd = 1.22$), and the RRT ($mean = 3.00, sd = 1.49$), $\chi^2(2,36) = 12.78, p = 0.001, W = 0.18$. Post-hoc pairwise comparisons found the PLT ($p = 0.008$) and the PSP ($p = 0.040$) were significantly more comfortable than the RRT, but not significantly different from each other ($p = 1.000$).  


\subsubsection{PSP was selected as most natural, but naturalness ratings were similarly low for all paths} \label{sec:natural}
The path that was selected as the most natural the most often was the PSP, chosen 15 times ($42\%$), followed by the PLT path, chosen 12 times ($33\%$), and finally the RRT was chosen as the most natural of the three paths the least often, nine times ($25\%$). However, a Friedman's test was run to compare the Likert response ratings of each of the paths from not natural at all (1) to very natural (6). There was no statistically significant difference between the PLT ($mean = 2.81, sd = 1.35$), the PSP ($mean = 3.08, sd = 1.25$), or the RRT ($mean = 2.86, sd = 1.52$), $\chi^2(2,36) = 1.17, p = 0.556, W = 0.02$.

\subsubsection{Preference was strongly associated with comfort}
A chi-square test of independence on the relationship between the path that was rated as preferred and the path that was rated as the most comfortable was significant, $\chi^2(4,36) = 11.18, p = 0.025, w = 1.14$. Due to the small sample size, a Fisher's exact test (two-sided) was also run for robustness and found to be significant, $p = 0.008$. People who rated the PSP as the most comfortable tended to prefer that path; however, people that preferred the PSP found the PLT path to be the most comfortable slightly more often. People that preferred the PLT path rated it as the most comfortable with only one exception; however, people that rated PLT as the most comfortable were evenly split on preferring the two Pareto paths. The crosstabulation tables of responses are available online at https://osf.io/et9gy. 

\subsubsection{Preference was strongly associated with naturalness}
The relationship between the path that was rated as preferred and the path that was rated as the most natural was also significant, $\chi^2(4,36) = 15.51, p = 0.004, w = 0.91$.  A Fisher's exact test was significant as well, $p = 0.003$. Thus, people who rated the PSP as the most natural tended to prefer that path, and people that rated PLT as the most natural tended to prefer that path over the others. 

\subsubsection{Naturalness was not associated with comfort}
A chi-square test of independence on the relationship between the path that was rated as the most natural and the path that was rated as the most comfortable was not significant, $\chi^2(4,36) = 2.59, p = 0.628, w = 0.85$. The Fisher's exact test was also not significant, $p = 0.587$. 

\subsubsection{PSP and RRT were rated as too close to the walls and objects more often than the PLT}
The responses to the question asking subjects to rate the distance between themselves and the walls or objects in the museum from 1 (too close) to 6 (too far away) were compared with a Friedman's test (the actual values for each path are listed in Tab.~\ref{tab:path_data}). There was a statistically significant difference between the paths, $\chi^2(2,36) = 36.95, p < 0.001, W = 0.51$ for the PLT ($mean = 3.72, sd = 0.78$), the PSP ($mean = 2.64, 1.15$), and the RRT ($mean = 2.03, sd = 1.00$). Post hoc pairwise comparisons found users perceived the RRT as significantly closer to the walls and objects than the PLT ($p < 0.001$), and the PSP as significantly closer to the walls and objects than the PLT ($p = 0.004$), but there was no difference between the PSP and the RRT ($p = 0.088$).

\subsubsection{PLT turns were rated to too fast more often than the RRT}
Subjects rated the turn speed for each path from 1 (too slow) to 6 (too fast) on a Likert scale question. The differences between the paths was significant, $\chi^2(2,36) = 12.31, p = 0.002, W = 0.17$. The PLT ($mean = 4.83, sd = 0.97$) was perceived as significantly faster ($p = 0.024$) than the RRT ($mean = 3.94, sd = 1.26$). However PSP ($mean = 4.61, sd = 1.15$) was not significantly different from the RRT ($p = 0.065$), nor were the two Pareto paths significantly different from each other ($p = 1.000$).


\begin{table*}[ht]
        \centering
        \small \caption{The numerical statistics of each path. 
        As the RRT is a smooth path, the number of turns cannot be defined.}
        \vspace{-1mm}
        \begin{tabular}{@{}cccccccc @{}}
         {} & {} & Mean fwd.~speed (m/s) & Mean turn speed (deg/s) & Duration (s) & Path length (m) & Min. wall dist. (m) & \# of turns  \vspace{0.5mm} \\ \hline \vspace{0.5mm} 
        {} & PLT &  1 & 90 &  76 & 72 & 1 & 2 \\ \hline \vspace{0.5mm} 
        {} & PSP & 1  & 90  &  67 & 62.6 & 0.4 & 4 \\ \hline \vspace{0.5mm} 
        {} & RRT & 0.52 & 18.8 &  129 & 67.2 & 0.3 & - 
  \end{tabular}
  \label{tab:path_data}
  \vspace{-4mm}
\end{table*}

\subsubsection{VR gaming frequency was not associated with the choice of preferred or comfortable paths}
To examine the relationship between VR gaming frequency and path selections, an open-ended question (``Have you used any VR systems previously and how many times?") was coded into three categories: \textit{never}, \textit{sometimes} (between one and nine times) and \textit{often} (10 times or more). One subject had to be excluded from the analyses for not having answered the question, thus the sample for these tests was 35 subjects. The association between VR gaming frequency and path preference was computed using crosstabulation and a Fisher's exact test (two-sided) to account for the small sample size. There was no statistically significant relationship between preferred path and VR gaming, $p = 0.854, w = 0.62.$ 
There was also no statistically significant relationship between the choice of most comfortable path and VR gaming frequency, $p = 0.710, w = 0.82.$ 
Finally, VR gaming frequency and choice of path that was the most natural was also compared using crosstabulation and a Fisher's two-sided exact test. The observed effect size for this association was moderate at $w = 0.41$, and power analysis revealed 71 subjects would have been required to achieve 80$\%$ power to detect an effect of this magnitude or greater. Thus, this experiment did not achieve sufficient power to detect an association between VR gaming frequency and choice of the most natural path. 

\begin{figure}[t]
    \centering
    \includegraphics[scale=0.55]{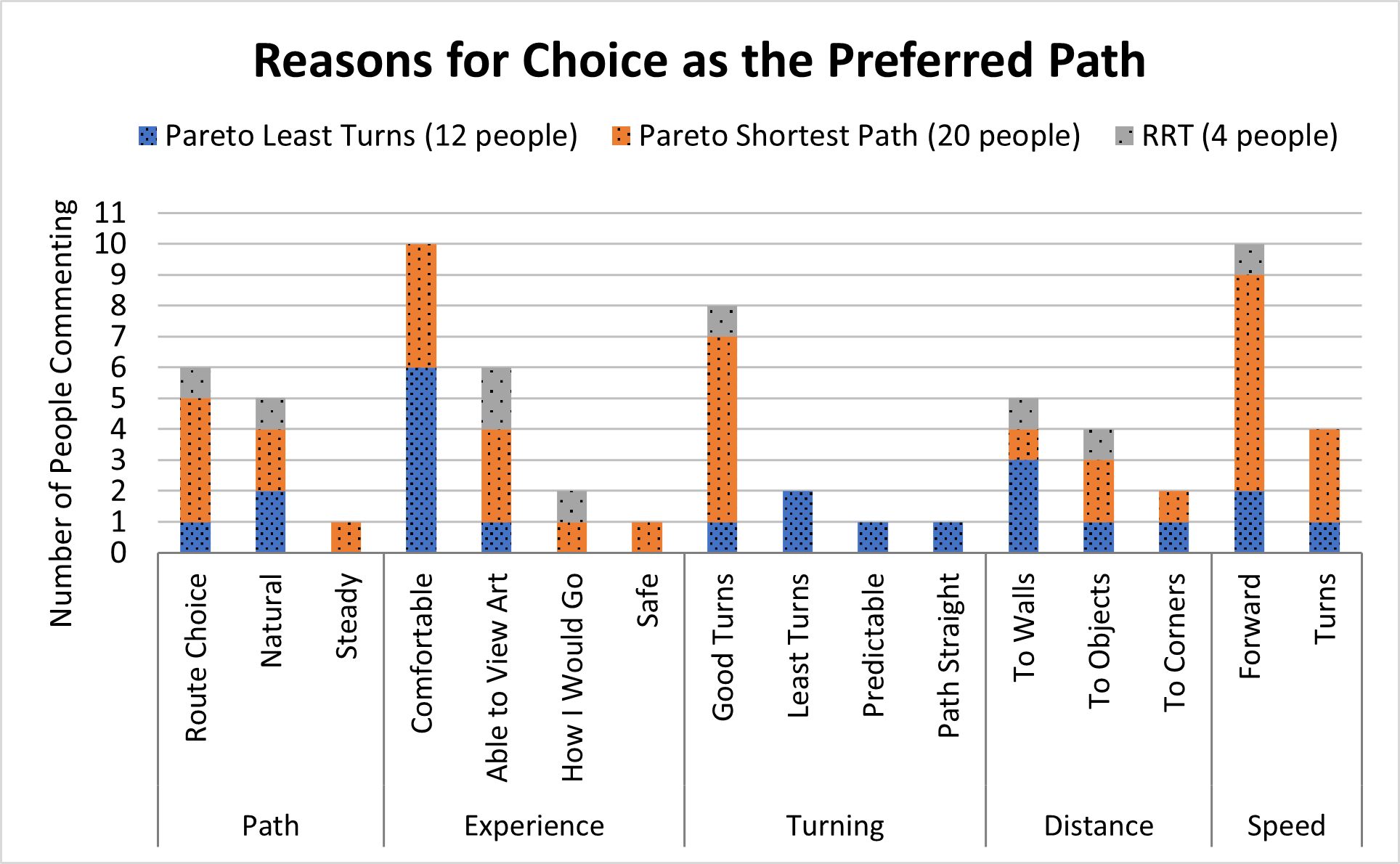} 
    \caption{Distribution of responses to the open-ended question asking subjects why they selected one of the three paths as the preferred path. 
    }
    \label{fig:preference} 
    \vspace{-4.4mm}
\end{figure}

\subsection{Qualitative Data}

\subsubsection{Comfort and forward speed influenced preference the most}
The responses to the open-ended question asking why subjects selected their preferred path can be seen in Fig.~\ref{fig:preference}. The greatest number of grouped comments (19, 28\%) regarded the experience on that path. The greatest number of individual comments was on the comfort, with six people describing the PLT as comfortable, four people for the PSP, and none for the RRT. The forward speed had a similar number of comments as comfort, with two people for the PLT, seven people for the PSP, and one person for the RRT.  

\subsubsection{Turns influenced comfort the most}\label{sec:turnspeed}
In response to the open-ended question regarding if paths were uncomfortable and why, 36$\%$ percent of the total number of comments were regarding the turns, which was the group of responses with the most comments for that question (59 comments on the turns, 35 on the path, 30 on the experience, 26 on the distance to objects, and 12 on the forward speed). The greatest number of individual comments regarded fast turns (11 people for PLT, six for PSP, two for RRT) followed by comments on being close to the walls (two people for PLT, eight for PSP, and eight for RRT). 

\begin{figure}[t]
    \centering
    \includegraphics[scale=0.55]{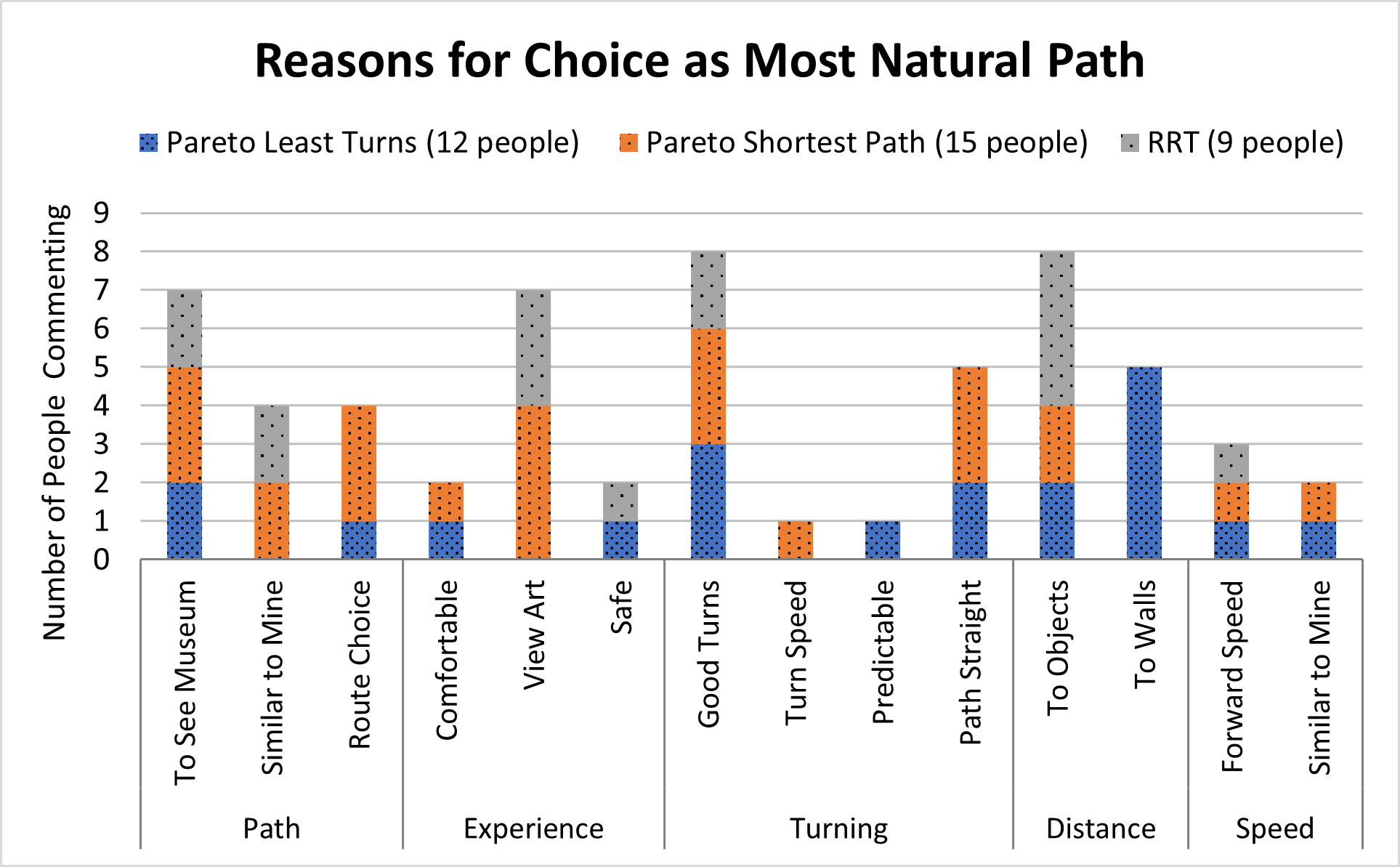} 
    \caption{Distribution of responses to the open-ended question asking subjects why they selected one of the three paths as the most natural. 
    }
    \label{fig:natural} 
    \vspace{-4.4mm}
\end{figure}

\subsubsection{The distance to objects and the turns influenced naturalness the most}
The group of reasons for selecting one of the three trajectories as the most natural was most often (15 total comments, 25\%) related to the path that was taken through the museum and the turns (see Fig.~\ref{fig:natural}). 
Individually, the greatest number of comments were on good turns (three people for PLT, three for PSP, two for RRT) and the distance to objects (two people for PLT, two for PSP, four for RRT). 

\subsubsection{Distance to walls and objects influenced comfort, naturalness, and preference}
In the responses of why subjects gave a certain Likert score for their rating regarding the distance between themselves and the walls and objects in the museum, the two most common responses for the PLT were that it kept a good distance (16 people, 44$\%$ of the sample) but also that it was too far away from the objects (10 people, 28$\%$). The PSP got too close to the corners (13 people, 36$\%$) and too close to the walls (10 people, 28$\%$). The RRT also got too close to the walls (18 people, 50$\%$) and too close to the corners (10 people, 28$\%$). 
The distance to walls, corners, and objects was also mentioned in the open-ended question responses regarding why a path was preferred (11 times), the most natural (13 times), or the most comfortable (13 times).

\subsubsection{Forward speed influenced comfort}\label{sec:forwardspeed}
The forward speed was mentioned several times as being a factor of significance for subjects. In the question regarding the reasons that a path was uncomfortable, six people (17\% of the sample) said the PLT and four people (11\%) said the PSP were too fast. However, on the question regarding which path was the most comfortable, three people (8\%) said PLT and one person (3\%) said the PSP because of their slow forward speed. The RRT was only mentioned as being uncomfortable once because of being too slow, though it was mentioned by a different person as a reason why they preferred that path.

\subsubsection{Descriptions of naturalness were contradictory}
Participants gave interesting, and at times conflicting, responses regarding the naturalness of certain paths. For instance, some subjects defined naturalness in terms of being circuitous, \textit{``I would visit one side first and then go to the other side of the museum, slowly exploring. So the first one} [PLT] \textit{was more natural"} and \textit{``Because this was about wondering [sic] around"} [RRT]. Other subjects described naturalness as more direct, \textit{``Walking through a corridor in straight line} [PSP] \textit{felt more natural than it was in the first path} [RRT] \textit{where the movement was from one corner towards another"} and \textit{``To me again, to go from point A to B, this path} [PSP] \textit{was the most natural, since it followed straight path [sic] without excessive turns. Not very natural, if you would like to explore the objects of the museum though"}. Participants also talked about naturalness in terms of the distance to walls and objects, \textit{``It felt most like a path that an actual human would take, getting closer to inspect the paintings"} [RRT] and \textit{``I think in real life I would move similarly to the 3rd path} [RRT]\textit{, because it would allow me to view the items better (closer). Also, it simulates the hesitation of a visitor when going in around a museum"}. Other subjects described naturalness in terms of the speed and turns, \textit{``the walking speed was closet [sic] to my own"} [PSP], \textit{``It feels stable due to the appropriate speed and less abrupt turns"} [PLT], and \textit{``This speed is more natural for a scenario or location such as this. However, the turns were a bit sharp and you feel like you're too close to the objects"} [RRT]. Finally, several described the most natural path as similar to their own, \textit{``I would have go [sic] that path if I were at the museum"} [PLT] and \textit{``Because I would have gone that path"} [PSP].

\subsubsection{Comfort influenced preference} When asked the open-ended question ``why?" regarding their reasoning for selecting a particular path as their preferred path, 10 participants (28$\%$) mentioned the word ``comfortable". For example, \textit{``Not to [sic] many turns. I feel comfortable"} [PLT] and \textit{``First path} [PLT] \textit{had sharp turns, second path} [RRT] \textit{was very inconvenient. Third path} [PSP] \textit{felt more pleasant in terms of speed and turns."} On a separate question regarding why a path was uncomfortable, the turns, speed, and distance to walls and objects were often mentioned, \textit{``Had a jerking sensation when turning. It seemed bit [sic] too quick. I noticed myself holding the chair tightly"} [PLT] and \textit{``it felt like the robot was like a roomba just bumping from walls to walls, not knowing clearly where to go. it didn't allow me to look around enough and that was frustrating"} [RRT].

\subsubsection{Naturalness influenced preference} When asked the open-ended question ``why?" regarding their reasoning for selecting a particular path as their preferred path, five participants (14$\%$) mentioned the word ``natural". For example, \textit{``It felt the most natural (moving closer to items/walls for better observations as opposed to going in the middle of the path) and would be closest to how I would go around in a museum in real life"} [RRT]. Some participants commented on both the comfort and naturalness; for example, \textit{``I felt more comfortable. The speed was almost natural to my speed when I walk although still a bit too fast for a place like a museum"} [PLT] and \textit{``It was more natural than the others and I have felt very comfortable, the robot was also far from the objects and walls"} [PLT].

\subsubsection{Naturalness and comfort conflicted} Though one person described their reason for selecting the PLT as the most natural \textit{``because it made me feel less sick than the others"}, others defined naturalness as being at the expense of comfort, \textit{``It tried hard being the most natural by getting closer to objects and cutting near corners but the experience wasn't comfortable and the turn's [sic] were slightly nauseating because they weren't smooth enough"} [RRT], and \textit{``Well it's subjective as I would love to take route which shows art naturally but I would't [sic] walk this close to wall. Maybe I would but it was uncomfortable in VR"} [PSP].

\section{DISCUSSION}
We set out to examine the interplay between the preferences, comfort, and perceived naturalness of robot motions for first-person experiences in VR telepresence. We found that both comfort and naturalness had strong associations with preference, but not with each other. The greatest number of comments in the open-ended question responses regarding preference were on the comfort of the path and the forward speed. The most salient aspect regarding comfort was in relation to the turns. Finally, the most influential aspects affecting naturalness were the distance to objects and the ability to see them, as well as the turns. 

Several aspects of the robot motions were mentioned frequently as the reason for a certain path to be preferred. In particular, the forward speed was listed by 10 people (28$\%$ of the sample). The forward speed was the same for the two Pareto paths, but the RRT was about half of their speed on average. Only one of the favorable comments on the forward speed was for the RRT, so the faster average speed of one meter per second was preferred. The turns were also one of the most influential aspects on preference, as well as comfort and naturalness. There was no significant difference in score on the Likert ratings of the turn speeds for the two Pareto paths, but the PLT was mentioned as having turns that were too large, too fast, and too sharp more often than the other paths. This could have been related to the fact that, although the speed of the turns on the two Pareto paths were the same, the PLT only had 90$^{\circ}$ turns, whereas the PSP had 45$^{\circ}$ and 90$^{\circ}$ turns. Despite the turns being reported as less comfortable on the PLT, the fact that there were the least amount of them likely contributed to the PLT being selected as the most comfortable of the three paths.  


Another striking result was the impact of closeness to corners, walls, and objects. Fear of colliding with walls, corners, or objects was specifically mentioned 17 times  on the open-ended questions regarding discomfort and distance, three for the PLT (minimum clearance 1 meter), eight for the PSP (0.4 meters), and six for the RRT (0.3 meters). Complaints about being too close to the corners (17 times for PSP, 12 for RRT) and objects (three times for PSP, three for RRT) were not mentioned regarding the path that was rated as most comfortable, the PLT. Even though the minimum distance criteria to walls and objects was actually violated for the most amount of time on the PLT path, the instances of extreme violation on the PSP and the RRT, though fewer, seemed to have a more significant impact on the users' perception of closeness. Whereas users commonly perceive distances as shorter in real world than in VR by approximately 10\% \cite{maruhn2019measuring}, which has been shown true in 360$^{\circ}$ videos as well \cite{el2020distance}, there is a known exception where the users perceived the distance the same in a virtual environment matching their real environment \cite{interrante2006distance}. 
We suggest that a minimum clearance of 0.4 meters should be taken into account when planning immersive telepresence robot paths.

\begin{figure}
  \centering
  \includegraphics[scale=0.4]{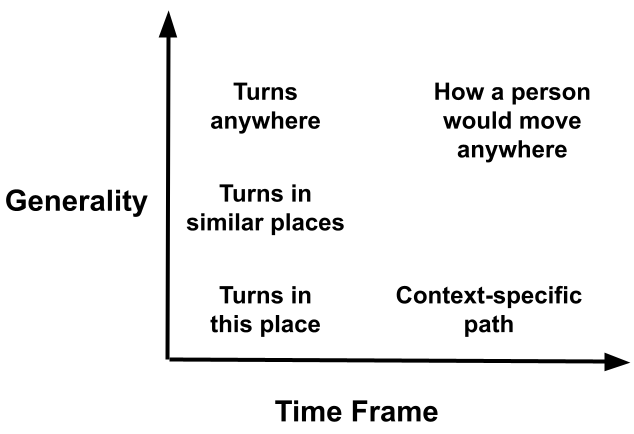}
  \caption{Conceptualization of the dimensions on which subjects consider movement to be natural. Horizontally is the time frame dimension. Things closer to the origin are more immediate (turns), whereas away from the origin, longer term, relates to things like the path. On the generality dimension, things closer to the origin are specific to a place (in this museum) whereas things away from the origin are more general (in any setting). 
  }
  \label{fig:axes}
  \vspace{-4mm}
\end{figure}

Although naturalness is a topic often mentioned as a goal in robotics publications, there is no widely accepted definition of naturalness, and it can mean different things to different people. To elucidate how naturalness is considered by people aboard a mobile robot, we asked participants open-ended questions regarding the naturalness of the robot's motions. We examined why people considered a path to be the most natural and found that there seem to be two dimensions (Fig.~\ref{fig:axes}) 
on which people interpreted the given definition of naturalness, ``how you would have moved through the museum if you were in control": time frame, and generality. On the time frame dimension, responses regarding either how the robot performed the turns (15 comments, 25\% of the total comments on naturalness) or how close to walls and objects the robot moved (13 comments, 22\%) can be considered to reflect something more immediate, while comments related to things further in time referred to aspects of the whole path through the museum (15 comments, 25\%). Regarding the generality dimension that describes either context-specific or more general behavior, subjects specifically mentioned the museum setting in seven answers spread across all paths, but some comments regarding turning, for example, could be considered as not specific to the context. Further discussion on naturalness can be found in our workshop paper \cite{Mimnaugh_Suomalainen_Becerra_Lozano_Murrieta-Cid_LaValle_2021}.

Finally, a number of participants said how well they could see the art influenced their preference and naturalness choices. Moreover, despite the wide range of experience using VR, from people that had never used an HMD to people who reported using one several times a month over an extended period of time, we did not detect a relationship between VR gaming frequency and choice of the most comfortable path. Therefore, in this case, context-specific issues regarding the turns and distances appear to have had a greater impact on comfort than issues that are traditionally considered when trying to reduce VR sickness, such as vection. This strengthens the intuition that Pareto-optimization or another kind of multi-objective optimization to create agreeable motions for a telepresence robot is needed, and simply trying to minimize VR sickness conventionally without considering other factors may not be the best approach.

\subsection{Limitations
}
There were a few limitations to our study. First, all three paths had the same maximum speed, but due to the nature of the RRT and the dynamics of a DDR when turning while continuing to move forward, the average speed of the RRT path was significantly lower. Consequently, the length of the video of the RRT path, which was almost twice as long as the videos of other two paths, likely had some effect on subjects' judgements regarding comfort as their exposure to the VR stimulus was longer. Second, the PLT path went though a narrow hallway without any art to judge the effect of an extended minimum distance violation, but then passed through the gallery at the end (see Fig.~\ref{fig:museum}). Thus, being further away from some of the statues in the gallery seemingly influenced participants' evaluation of the paths. However, 
other aspects, such as the forward speed and the turns, appear to have had a greater impact. 
Counterbalancing the placement of the art could address this issue in future experiments. Additionally, people, in general, have different behavioral patterns when visiting a museum, which we did not take into account \cite{kuflik2012analysis}. Third, this study had a small number of subjects, which limited the external validity. Nevertheless, the strong effect sizes were sufficient in all but one of the tests to achieve greater than 80$\%$ power to detect effects of those magnitudes or greater. Fourth, the study was run in simulation, which also limits the external validity. Our subsequent work will test with a physical robot on our university campus to compare sim-to-real results. Fifth, there was some ambiguity in the responses to the open-ended questions, which were at times difficult to code or interpret as many subjects were not native English speakers. This could have also influenced our results if the subjects interpreted the meanings of the English words used in the questionnaires in different ways. Finally, we were only able to test aspects of the paths for one type of path per condition. As there is a limit on how much a human subject can be exposed to a virtual environment before either they begin to feel very uncomfortable or they begin to adapt (which would confound the results), we were limited to only presenting a few paths to each subject. In the future, we plan to test additional variations for speeds and distances to objects to further elucidate subject preferences regarding these aspects.


\section{CONCLUSIONS}\label{sec:con}

Our work examined the interplay between preferences, comfort, and naturalness for users immersed in a telepresence robot through a head-mounted display. To our knowledge, this is the first work to investigate user path preferences for first-person autonomous telepresence robot motions viewed with an HMD. We hope that this work can stimulate further interest in designing immersive robotic telepresence experiences that address user experience so that this technology can be more effective and more widely used in the future.






\section*{ACKNOWLEDGMENT}

The authors would like to thank Evan Center, Ba\c{s}ak Sak\c{c}ak, Matti Pouke, Vadim Kulikov, Mattia Racca, and Michael Mimnaugh for feedback on the manuscript. They would also like to thank the research participants.


\bibliographystyle{ieeetr}
\bibliography{references}


\end{document}